# A tool set for the quick and efficient exploration of large document collections

**Camelia Ignat, Ralf Steinberger, Bruno Pouliquen & Tomaž Erjavec**

Institute for the Protection and Security of the Citizen

Joint Research Centre, European Commission

Via Fermi 1, Ispra 21020 (VA), Italy

E-mail: Firstname.Lastname@jrc.it

*Abstract:*

*We are presenting a set of multilingual text analysis tools that can help analysts in any field to explore large document collections quickly in order to determine whether the documents contain information of interest, and to find the relevant text passages. The automatic tool, which currently exists as a fully functional prototype, is expected to be particularly useful when users repeatedly have to sieve through large collections of documents such as those downloaded automatically from the internet. The proposed system takes a whole document collection as input, carries out some automatic analysis tasks, annotates the texts with the generated meta-information, stores the meta-information in a database, and provides the users with an interface that allows them to search and view the most pertinent text passages. In the first step, named entities (names of people, organisations and places) are recognised and stored. Then, highly similar documents are grouped into clusters of documents that are likely to report about the same event or subject domain. In the third step, domain-related terms are detected and extracted from each cluster of documents. Then, a zoomable map is generated for each of the document clusters to show the geographic coverage of the collected documents. Additionally, names and specialist terminology found in the document cluster are listed. Hyperlinks allow the users to jump to the text passage where they were found, and also to read the individual documents that are part of the cluster. All clusters of the document collection are made available in an html file that can be viewed with a standard web browser. When the system is used on a regular basis, it builds up a historical database that contains information on which names have been mentioned together with which other names or places, and users can query this database to retrieve information extracted in the past.*

## 1. Introduction

Most information-seekers and analysts feel that the so-called *information overflow* is a major restriction to their work. Although a lot of useful information is freely available on the internet and from other sources, it gets increasingly difficult to *find* the relevant information. Search engines are getting better at identifying the most relevant documents on the internet, but users still need to look through many documents to find the piece of information they need.

We are proposing a solution that helps information seekers to sieve through large collections of documents quickly and to speed up the process of finding what they are looking for. Assuming that users already have a large document collection to look through, the solution consists of automatic text analysis software that firstly clusters all related documents into groups, and that then extracts information from each cluster and presents it to the users in an organised and intuitive manner. For each cluster, the extracted information consists of: (a) a list of keywords that indicate the approximate contents of the document cluster, (b) a list of geographical places mentioned in the cluster, lists of (c) names of people and (d) of organisations, and (e) a list of the user interest-specific spe-cialist terms found in the cluster. In principle, further information such as lists of all dates, addresses, telephone numbers, etc. could be extracted and displayed, as well. Figure 1 shows how extracted information for a given cluster of documents can be presented to the user.

The proposed system currently exists as a prototype that is specialised in *news* analysis, i.e. on the processing of thousands of news articles every day. Part of the system can be tested at the address http://press.jrc.it/NewsExplorer.

The application consists of many individual components which cannot all be described here in detail. The purpose of this article is to show the potential of such a system, without going into too many technical details and without always comparing the individual applications with the state-of-the-art. Instead, we will refer to some of our more specialised publications where the technology used is described in more detail.

Following this introduction, we first describe ways of gathering potentially user-relevant documents and of preparing them for an automatic analysis, including the identification of document duplicates and near-duplicates (section 2). We then dedicate some space to the analysis tasks of clustering related articles (sect. 3) and of extracting keywords, places, names and terms from them (sect. 4). Section 5 and 6 are dedicated to the visualisation and presentation of the extracted text data, as well as to ways of helping users cope with foreign language text collections. Section 7 gives information on the usability status of the prototype software.

## 2. Gathering and pre-processing the document collection

Information-seeking users may either already have collections of documents that they need to look through, or they may use automatic document gathering tools that automatically collect documents that are potentially relevant. Commercial web crawlers that download documents from specific web sites exist, but it is also possible to develop a tool that exploits the power of existing web search engines like Google by downloading all the documents found by the search engine. Furthermore, there are crawlers that specialise in specific document types such as news. An example of such a system is the *Europe Media Monitor* EMM (http://emm.jrc.org; Best et al. 2002), which monitors about 20000 news in thirty languages a day.

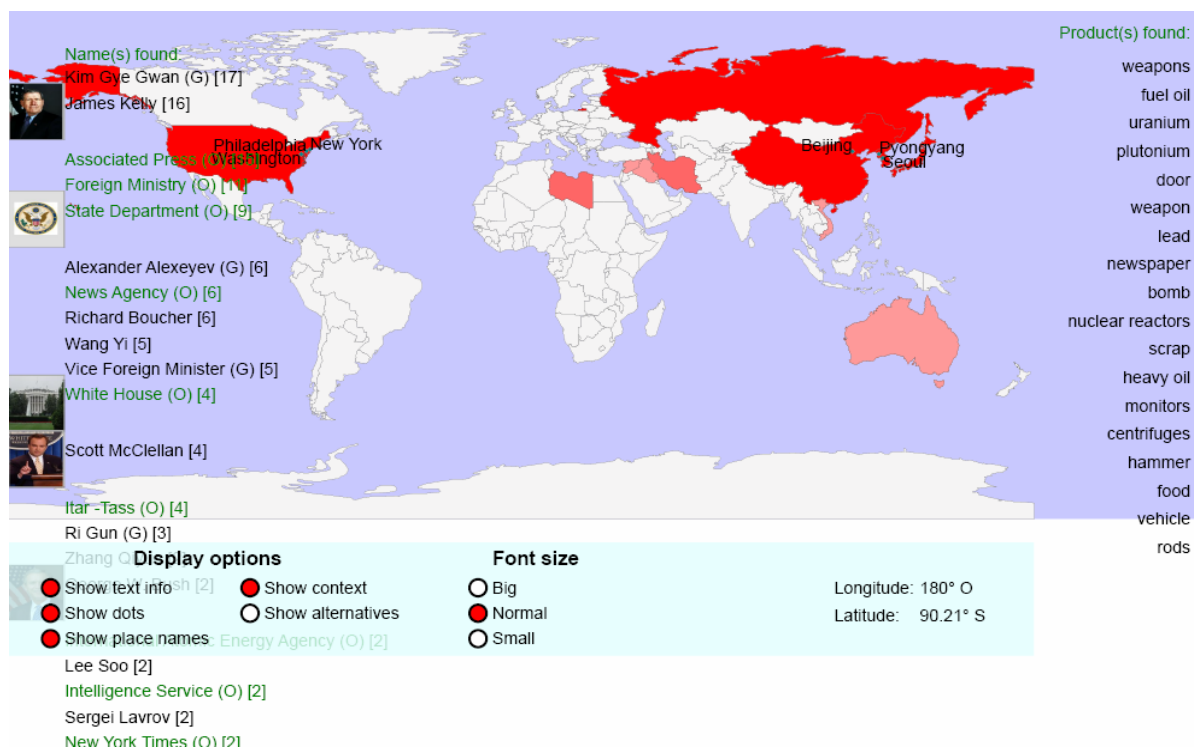

**Figure 1.** Automatically generated map showing the geographical coverage of one news cluster on North Korea, consisting of 26 articles from 10 different newspapers. The map also shows specialist terms as well as the names of people and organisations automatically extracted from the text collection. Hyperlinks on the places and on the terms allow users to jump directly to the relevant text passages, while a click on any of the names leads to a database containing additional information on the people and organisation.

## 2.1. Document format conversion

In which ever way the collection is being compiled, it is likely that the documents are in different file formats, such as HTML, PDF, MS-Word, postscript, plain text, etc. Automatic text processing tools require that the text be available in machine-readable format such as plain text or, ideally, in a more structured format such as XML. Tools that take proprietary file formats and convert them to plain text or XML are commercially available. For news analysis, the standardised XML subtype RSS is appropriate, as it is widely used (Hammersley 2003).

## 2.2. Identification of document duplicates and near-duplicates

There is a certain likelihood that the document collection contains several copies of the same or very similar documents. At the *Joint Research Centre*, we are using a duplicate identification software tool that has been adapted from software developed at the German *Institut für Deutsche Sprache* in Mannheim. The tool splits every document of the collection into word pentagrams (i.e. five consecutive words) and checks, for each pair of documents, whether some of the pentagrams are identical. If more than a certain percentage of all pentagrams are identical (e.g. 50% or more), this can be flagged to the user or the duplicate documents can automatically be removed, depending on the users' preferences. The tool thus identifies whether entire texts or some text passages are *identical* between two documents. By doing this, multiple reading of the same text can be avoided to save time or to avoid duplicate counting of the phenomena the information seeker is looking for.

## 3. Grouping of documents according to similarity

Several non-identical documents are likely to describe the same contents. In news analysis, for instance, we found that a main event (such as the Madrid Bombing in March 2004 or the death of Pope Jean-Paul II in April 2005) can cause the publication of over one hundred news articles per day in English alone. Some of them are partial duplicates (e.g. all news articles based on the same news wire), others simply talk about the same things. It is useful for the information seeker to know which documents are in such a way related in order to either discard them altogether (if they are not of interest) or to concentrate their effort on them.

The grouping of related documents can be automated in a statistical process called *clustering*. In the clustering process, software calculates the similarity between each pair of documents. All those document pairs or document groups that are more similar than a certain threshold can then be grouped into a single cluster (see Pouliquen et al. 2004b). The following sections explain how we represent documents by their keywords and the countries mentioned in them, and how we then calculate the similarity between the documents.

## 3.1. Keyword identification for document representation

To produce a meaningful and machine-readable representation of each document, we represent each text by a ranked list of its most pertinent words (the *keywords*) and their relative importance (the *keyness* of the keywords). The approach to produce keyword lists for a new document exploits the fact that words have an average frequency in texts. For instance, the word 'the' occurs about 60000 times per million words, while the words 'table' and 'xenophobia' occur about 200 and 1 times per million, respectively. If a word occurs considerably more frequently in a given text than it occurs on average (normalised by the total number of words), then this word is likely to be of some importance, i.e. it is a keyword. *Keywords* are thus those words that occur outstandingly frequently in the given text. *Keyness* is the degree with which they are outstandingly frequent. Both keywords and their keyness can be calculated with statistical formulae such as TF.IDF or the *log-likelihood* or *chi-square* tests (See Kilgarriff 1996). At the JRC, we use the *log*-likelihood formula to calculate which words are keywords, as it seems to perform best for our kind of data (Kilgarriff 1996). Table 1 shows the top of the automatically produced list of keywords for the document you are currently reading.

| Keyword | Keyness |
|---|---|
| cluster | 714 |
| name | 361 |
| jrc | 334 |
| keyword | 320 |
| tool | 317 |
| pouliquen | 290 |
| place_name | 264 |
| figure | 264 |
| 2004 | 238 |
| language | 224 |
| keyness | 187 |
| 2003 | 187 |
| ignat | 169 |
| 2005 | 162 |
| collection | 160 |
| multilingual | 155 |
| analysis | 145 |
| steinberger | 141 |
| table | 137 |
| display | 128 |
| word | 125 |
| automatically | 124 |
| news | 113 |
| … | … |

**Table 1.** Sample document representation of the document you are currently reading, i.e. a list of keywords and their keyness.

## 3.2. Particular consideration of geographical information

In order to distinguish news about similar events (such as election campaigns or reports on earth quakes) in different countries, we add geographical information to the list of keywords. We do this by first identifying references to places in each article (see section 4.2), and by adding to the country score of the article each time we find a reference to a city or town of that country. As some countries generally are being talked about much more than others, we furthermore normalise the country score by comparing the count with average counts, using again the log-likelihood formula. This means that if a document mentions 'France' and the cities of 'Paris', 'Lyon' and 'Tours', this is counted as having four references to the country 'France', and this number is normalised by the average number of references to 'France' in texts of a similar type (e.g. news). The list of a document's keywords is thus enhanced by the country ISO code 'FR' and its keyness value. The resulting list of keywords and country codes, together with their keyness values, can be seen as a vector that represents the document.

## 3.3. Document similarity calculation and clustering

Similarity calculation can be based on lexical overlap or other measures. Lexical overlap means that two documents are judged to be similar if they are made up of the same or similar words. This means that the more words they have in common, the more they are judged to be similar. High-frequency words that occur in all sorts of documents such as *the, of, for, from*, etc. (*stop words*) are usually not considered in the comparison. At the JRC, we base our similarity calculation on the vectors produced in the steps described in sections 3.1 and 3.2.

The similarity between two documents can then be calculated as the *cosine* between the vector representations of the two documents. The cosine expresses the similarity or distance of two vectors: If the cosine is 1, the vectors (and thus the documents) are identical. If the cosine is close to zero, the vectors are very dissimilar.

The grouping of related documents can be automated in a statistical process. In this clustering process, software calculates the similarity between each pair of documents. Documents that are more similar than a certain threshold can then be grouped into one cluster. For the analysis of collections of news articles, we found empirically that the threshold of 0.5 produces meaningful results (Pouliquen et al., 2004b).

For each cluster, the most typical document is identified, i.e. the document that is most similar to the centroid in the document cluster's vector space. The title of this centroid article is used as the name for the cluster.

## 4. Extracting information from each cluster of documents

In our current application, each cluster of related texts is henceforth treated as if it were a single (meta-) document about a certain event or story. For this meta-document, we automatically extract and present information. We currently identify keywords (section 4.1), information about geographical references (4.2), and information about references to people and organisations (4.3). Where available, we also search for the occurrence of terms from a user-controlled term list in the document (4.4). We plan to extract and display further information in the future, including dates, products, etc.

### 4.1. Keywords

For each cluster, we identify and display the main keywords in order to give users an overview of the contents of the cluster. In order to calculate the cluster's keywords, we take as input the keywords of each individual document, which were calculated in the process described in section 3.1. We then calculate the average keyness of each keyword across all individual articles to come up with the averaged keyness (*group averaging*). The keywords with the highest averaged keyness are then displayed to the users.

### 4.2. Geographical references

A geo-coding tool automatically recognises references to countries, cities, towns and villages. Rules and heuristics are used to distinguish place names from homographic person names and normal words (there are places called 'And' 'Split', 'Bush', 'Victoria', etc.), and to decide which of several homographic place names is being talked about in a text (world-wide, there are 45 places called Paris, 18 places called 'Birmingham', etc.). The result is a list of geographical place names found in the (meta)-document, together with information on the cor-responding country and with exact latitude and longitude information. A separate map generation tool produces zoomable interactive SVG maps (Eisenberg 2002) in which the countries and cities mentioned are shown (see Figures 1 and 2). Colour intensity is used to indicate the number of hits per country. When users point the mouse on a highlighted place or country, one-line of left and right hand side contexts of the place name are shown (similarly to Figure 3 for terms). The idea of this application was to let users investigate quickly in which context the place name was mentioned. The details of this application are described in Pouliquen et al. (2004a) and Ignat et al. (2003).

### 4.3. Person and organisation names

A name recognition tool also scans each (meta)-document to identify references to people and organisations. The current version of the software was tuned for news analysis. It identifies uppercase words in text as names either if they are already known names stored in a database, of if the context makes it clear that they are names. These contextual patterns are expressions that can be titles (Professor, President, Lithuanian Minister for Foreign Affairs), professions (painter, artist) or other trigger words ('has said', ). For English alone, the list contains about 1300 manually compiled and empirically derived context patterns. In one year of news analysis, we have found about 125000 distinct names. Every day, an average of 500 new names is added to the database.

In the current version of the JRC tool for the exploration of document collections, all names found in a given cluster are listed at the left hand side of the map showing the geographical coverage of the same document cluster (see Figure 1). Additionally, the trigger words that helped to identify the name can be displayed next to the name (e.g. 'US Secretary of State' or 'Iranian religious leader'). This is deemed useful to provide additional information on the person identified.

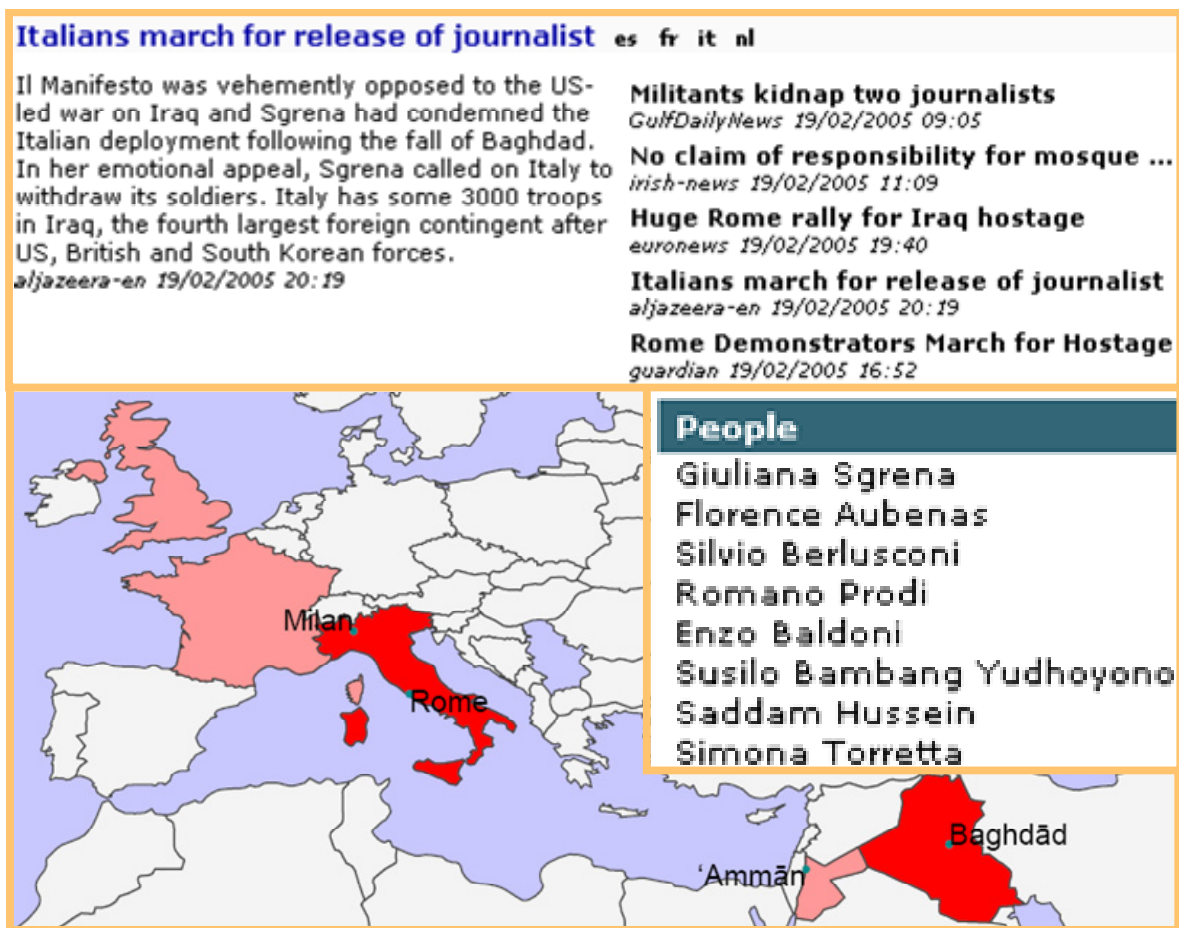

**Figure 2.** English news cluster with automatically identified links to related news clusters in Spanish, French, Italian and Dutch. Hyperlinks on the titles allow users to read the individual articles, links on names allow them to find out more about the persons, all news articles in which they are mentioned, etc.

**Figure 3.** All occurrences of the term 'plutonium' found in a cluster of news articles, and their context.

### 4.3.1. Identifying name variants

As the documents in a cluster normally come from different sources, the same name is often written in different ways (see Table 2). This is particularly true for Arabic and Russian names, as there are various transcription standards. For some names, we found over sixty variants in one year of news analysis. As it is useful to combine the variants under one single alias, or even a numerical person identifier, we automatically launch an approximate matching tool for all names found in a cluster. This tool splits each name into all possible sets of consecutive two-letter and three-letter groups (bigrams and trigrams, more generally *n-grams*) and then calculates the similarity between the n-grams of two names, using the cosine formula. When the overlap between two names is above a certain threshold, i.e. when two names have a large percentage of n-grams in common, then they are flagged as being variants of the same name. In order to be able to handle languages with different alphabets (Arabic and Russian), additional algorithms have been developed. For details on the name recognition and the approximate name matching work, see Pouliquen et al. (2005). The automatically identified variants are stored in a database to build up a list of all variants for the same name that have been found over time. Furthermore, all articles and clusters mentioning any of the variants are indexed with the numerical name identifier so that users can search the database for all articles about a certain person, independently of how the name was spelled. As the approximate matching sometimes misses variants, or it maps two names that are not the same, the automatically achieved results can be checked interactively and they can be corrected.

### 4.3.2. Identifying related persons

The JRC system keeps track of all person and organisation names mentioned in the same clusters. When names are mentioned repeatedly over time, a network of related persons and organisations thus develops, because some names will co-occur more often in the same clusters than others. The system learns over time, as it is updated daily. For a given person, users can thus ask the system to display lists of those persons that are most frequently mentioned in the same clusters. A more sophisticated weighting technique also allows to display mostly those names that are specifically associated to a given person, i.e. those that mainly occur with this person and not so frequently with other persons. For details, see Pouliquen et al. (2005).

### 4.3.3. Finding information on the internet

As information on some people is available in online encyclopaedias on the internet (e.g. http://en.wikipedia.org), we automatically check for each name identified whether information about this person is available. If it is, we store the URL of this page so that we can make it available to the user (see Table 2). That way, users can find additional information about a person with a simple click.

Mohammed ElBaradei
Mohamed El Baradei
Muhammad al-Baradai
Mohammed al-Baradei
Mohamed al-Baradei
Mohammed El Baradei
Mohamed El-Baradei
Mohammed el-Baradei
Mohamed el-Baradei
Mohamed el Baradei
Mohamed ElBaradei
Mohammed el Baradei

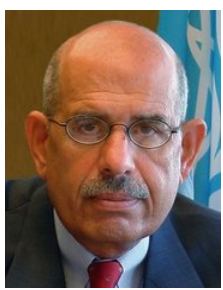

http://de.wikipedia.org/wiki/Mohammed_el-Baradei
http://en.wikipedia.org/wiki/Mohamed_ElBaradei
http://fr.wikipedia.org/wiki/Mohamed_ElBaradei
http://it.wikipedia.org/wiki/Mohamed_ElBaradei

**Table 2.** Name variants found in many different news articles and automatically identified as belonging to the person *Mohammed ElBaradei*. List of Wikipedia encyclopaedia entries automatically found and displayed to give users quick access to information about a person.

```
nuclear [39]
weapons of mass destruction [12]
uranium [7] (natural uranium|uranium)
plutonium [7] (plutonium)
disarmament [6]
missiles [5]
iaea [4]
missile [4]
proliferation [3]
atomic [2]
biological [1]
reprocessing [1]
enrichment [1]
nuke [1]
scud [1]
```

**Table 3.** List of specialist terms found in a cluster of six news articles, and their frequency.

### 4.4. Specialist terms

Depending on their job-specific interests, users may be looking for different terms in texts, such as nuclear terminology, names of diseases, financial terms, etc. If they can see, for a given cluster of tens of documents, all of their terms of interest that occur anywhere in the cluster, they can quickly decide whether the document cluster is of interest, what aspect of their work is covered by the group of documents, etc.

The current system therefore allows users to provide term lists with tens, hundreds or even thousands of terms of their field. The JRC tool then searches each cluster automatically for occurrences of any of the terms and lists them for the user (see Table 3). When clicking on a term in the term lists, a separate window opens to show the complete list of contexts (see Figure 3). Users can thus see the context of each of the terms without having to open the file in which the term occurred. It is planned to add a further hyperlink allowing the users to jump directly to a certain document in case the small context looks interesting, to allow a further exploration of the document.

**Semi-automatic creation of domain-specific term lists**

For the automatic display of the user-relevant terms found in a given cluster, long lists of potentially relevant terms are required. Users do not normally have such lists available, and potentially existing domain glossaries may not be suitable for the task as they often only contain the standardised version of terms, while other versions (synonyms or alternative spellings) may exist and may be used in real-life texts.

In order to quickly compile large lists of domain-specific terms as they occur in real-life texts, we use tools that automatically extract all potential terms from specialist documents provided by the users. These terms can then be ranked by domain specificity, using statistical methods. Users can then simply look through the resulting term lists and choose those words or word groups that they would like to be highlighted in texts. It is up to the users to decide whether they are interested in having only highly specific terms identified in a document cluster, or whether they also want to know about more general terms. The term lists in Figure 1 and Table 3 also contain rather general terms such as 'rods' and 'monitors'.

The terms can additionally be marked as belonging to different sub-areas, and this additional information can be displayed together with the terms. In the field of *Nuclear Safeguards*, for instance, terms could be divided according to the part of the nuclear fuel cycle they relate to. Enrichment-related terminology, for instance, could then be identified even more quickly.

It is our mid-term aim to provide users with term lists from various fields of interest (e.g. medicine, finance, computer science, etc.), from which they can choose. Once this aim is achieved, they can simply tick subject area boxes and the terms from the chosen fields will be displayed.

### 4.5. Language-specific issues

It is relatively easy to detect the occurrence of a term or of a place name in text if the word form found in the text is the same as the word form found in the list of terms or place names. However, special means need to be used, for instance, to find the plural form of a term in text (e.g. *centrifuges*) if the term list contains the singular term (*centrifuge*). For the English language, the problem is reduced to plural and to genitive (*centrifuge's*) forms. In more highly inflected languages such as those of the Slavic or Finno-Ugric language families, the number of word variants can be rather large. In Slovene, for example, the noun *centrifuga* can appear in any of the following forms: *centrifuga, centrifugi, centrifugo, centrifug, centrifugama, centrifugah, centrifugam, centrifugami.* In Russian, the following word forms for the equivalence of centrifuge can be found: центрифуга, центрифуги, центрифуге, центрифугой, центрифугу, центрифуги, центрифугам, центрифуг, центрифугами, центрифугах.

This problem can be solved by using morphological processing tools such as lemmatisers, which analyse the inflected word form and re-turn the uninflected or neutral dictionary form of the term. However, such linguistic tools are not available for all languages, and purchasing them for a large number of languages is rather expensive. At the JRC, we therefore tackle the problem by using simple heuristics and exhaustive lists of suffixes and their combination with the word stems. In Russian, for example, the suffix list for the stem 'центрифуг-' (*centrifuge*) contains the suffixes: 'а|и|е|е|ой|у|ам|ами|ах'. These lists are used to generate all term variations so that the term will be successfully found in text. For details, see Steinberger et al. (2004).

## 5. Organisation and visualisation

The results of the analysis of the whole document collection is a list of document clusters, where additional information is available for each cluster (see Figure 2). The JRC's system orders the clusters by size (largest first) and displays all the extracted information: cluster title, map showing the geographical coverage of the documents in this cluster, and the lists of names, organisations and terms. Alternatively, related clusters can be displayed together, independently of their size. The users can thus start exploring the document collection by clusters. They can decide whether to sieve through the collection by size, or whether they first want to identify those clusters that seem most promising because they contain many relevant terms or names, etc.

When the users click on any of the names, they will see a page showing the name, the name variants found, the list of encyclopaedia entries found (if any, see Table 2), a list of all articles and of all clusters mentioning this person, and a list of related names and organisations (see section 4.3.2). Users can thus explore and navigate the document collection, for instance by querying the database for all news clusters in which a person has been mentioned. They can find clusters mentioning a certain keyword or country, etc., and they can ask for further related clusters.

## 6. Overcoming the language barrier

A bottleneck in information search is the fact that some of the texts are written in languages the information seeker does not understand. While this problem cannot be solved completely, some tools and methods can at least be of some help. We suggest two strategies: exploit multilingual gazetteers, nomenclatures, term lists and thesauri for cross-lingual information display, or represent and display the textual information in a language-independent way. These methods can also be used to detect the similarity of documents written in different languages so that users can explore multilingual document collections more conveniently and efficiently. For a more detailed discussion of the methods presented here, see Steinberger et al. (2004).

### 6.1. Cross-lingual glossing

When information about the extracted term or place name is available in other languages, this information can be displayed inside the text (see Figure 4). For instance, when term lists are available in English and German, English-speaking users can ask the system to identify and highlight the specialist terms found in German text. Additionally, the English translation can be displayed so that the users see in their own language which terms were found in documents written in another language. Figure 4 is an example of automatic detection of place names, and their highlighting and cross-lingual glossing.

### 6.2. Language-independent representation of textual information

Some textual information can be normalised and represented in a language-neutral way, including date and currency expressions and geographical information. For the representation of references to geographical places, maps and latitude-longitude information are language-neutral (see Figures 1 and 3). Dates written in various formats and in various languages can be represented using a numerical standard (see Ignat et al. 2003). That way, a foreign language text string such as *petnajstega novembra dvatisočpet* can be displayed as 15-11-2005.

The same could be done for currency expressions, telephone numbers, number plates, etc. Such normalised expressions could be highlighted and displayed in the text, similar to the place names in Figure 4.

**Figure 4.** Place names identified and highlighted in Bulgarian, Czech and Estonian texts. Cross-lingual glossing in English.

### 6.3. Linking similar texts across languages

In news analysis, users are often interested in reading how the same subject or event was discussed by the media in other countries. Therefore, we developed a means to automatically show which clusters in various languages talk about the same story or event. For each given news cluster, the system produces a representation that can be compared to the representations in the other languages. This representation consists of a ranked list of classes from the multilingual thesaurus Eurovoc (Eurovoc 1995), of a ranked list of countries (derived from the country and city names mentioned), of a list of person names automatically identified, and of a list of keywords. The latter is not language-independent, but it benefits from the fact that some names, numbers and cognates are identical even across languages. The complex cluster representation vectors can then be compared so that all clusters being more similar than a certain threshold are identified as being related. For details, see Pouliquen et al. (2004b).

### 7. Usability status of the software / Future work

The various software tools mentioned in this paper have reached different degrees of maturity. Some of them have been integrated into one complex application so that the results can be browsed in the EMM *News Explorer* (http://press.jrc.it/NewsExplorer), which is currently available in the eight languages English, German, French, Spanish, Italian, Dutch, Slovene and Estonian. Other tools have no interface and need to be combined individually by programmers in order to perform a full analysis of a given text collection. None of the tools is currently set up to be installed on computers outside the JRC. The name recognition patterns are tuned to perform well on news articles, where persons are usually referred to with their title (e.g. *Foreign Minister*) and where persons are often referred to (at least once) by their first and last names. The system in its current state would therefore not work very well on different text types such as scientific publications, where first names and titles are not used.

It is planned to integrate a number of functionalities into a standalone version that can be installed at user sites and that can be used directly by various analysts. This means that automatic document gathering tools like crawlers and search engine result scrapers need to be integrated with document format converters and the whole suite of text analysis tools. Furthermore, an intuitive user interface needs to be prepared, and tools need to be more integrated and robust. We furthermore intend to extend the functionality to more languages and to make the name recognition tools more robust with respect to text types other than news. Over time, we intend to integrate text analysis tools that are not currently integrated (e.g. date recognition) or that will be developed in the future (e.g. recognition of currency expressions). Another plan is to prepare multilingual specialist term lists for various subject areas so that users will be able to select those subject areas for which they want the system to recognise and display terms in their texts, as shown in Figure 1.

### 8. Acknowledgement

Many people have contributed to developing the tool set described in this paper and to developing and evaluating the language-specific resources for various languages. We would like to thank Louis-Victor Bril, Joao Gonçalves and the other colleagues from the JRC's *Nuclear Safeguards and Non-Proliferation Unit*, who helped us specify the functionality of text analysis tools for analysts. We thank Marc Kupietz from the *Institut für Deutsche Sprache* for contributing the document duplicate identification tool and Peeter Päll from the *Institute of Estonian Language* for contributing his multilingual lists of place names. We would like to thank Marco Kimler for his refinement of the geo-coding tool, and Emilia Käsper, Ippolita Valerio, Tom de Groeve, Victoria Fernandez Mera, Tomaž Erjavec, Christian Gold, Irina Temnikova, Jan Žižka, Helen Salak, Wajdi Zaghouani, Anna Widiger and Pinar Özden for their help in creating language-specific resources for Estonian, Italian, Dutch, Spanish, Slovene, German, Bulgarian, Russian, Czech, Farsi, Arabic, Russian and Turkish. We would particularly like to thank Clive Best, Erik van der Goot, Ken Blackler and Teofilo Garcia from the JRC's *Web Technology* team for providing us with the multilingual news collection to develop and test many of the applications described here. We finally thank our former colleague Johan Hagman for introducing us to the methods and usefulness of cluster analysis.